\documentclass[10pt,twocolumn,letterpaper]{article}

\usepackage[pagenumbers]{cvpr} 
\usepackage{color}
\usepackage{pifont}
\usepackage[dvipsnames]{xcolor}

\usepackage{float}
\definecolor{cvprblue}{rgb}{0.21,0.49,0.74}
\usepackage[pagebackref,breaklinks,colorlinks,citecolor=cvprblue]{hyperref}
\usepackage{makecell}
\usepackage{footnote}
\usepackage{threeparttable}
\def\paperID{****} 
\def\confName{CVPR}
\def\confYear{2024}

\title{ChebMixer: Efficient Graph Representation Learning with MLP Mixer}

\author{Xiaoyan Kui \\
    Central South University
    \and 
    Haonan Yan \\
    Central South University
    \and Qinsong Li \\
    Central South University
    \and Liming Chen \\
    Ecole Centrale de Lyon
    \and Beiji Zou\\
    Central South University}

\usepackage{tabularray}
\usepackage{algorithm}
\usepackage{algorithmic}
\begin{document}

\begin{minipage}{\textwidth}\ \centering
		This work has been submitted to the IEEE for possible publication. Copyright may be transferred without notice, after which this version may no longer be accessible.
\end{minipage}

\maketitle

\begin{abstract}
Graph neural networks have achieved remarkable success in learning graph representations, especially graph Transformer, which has recently shown superior performance on various graph mining tasks. However, graph Transformer generally treats nodes as tokens, which results in quadratic complexity regarding the number of nodes during self-attention computation. The graph MLP Mixer addresses this challenge by using the efficient MLP Mixer technique from computer vision. However, the time-consuming process of extracting graph tokens limits its performance. In this paper, we present a novel architecture named ChebMixer, a newly graph MLP Mixer that uses fast Chebyshev polynomials-based spectral filtering to extract a sequence of tokens. Firstly, we produce multiscale representations of graph nodes via fast Chebyshev polynomial-based spectral filtering. Next, we consider each node's multiscale representations as a sequence of tokens and refine the node representation with an effective MLP Mixer. Finally, we aggregate the multiscale representations of nodes through Chebyshev interpolation. Owing to the powerful representation capabilities and fast computational properties of MLP Mixer, we can quickly extract more informative node representations to improve the performance of downstream tasks. The experimental results prove our significant improvements in a variety of scenarios ranging from graph node classification to medical image segmentation.

\end{abstract}

\begin{figure}[t]
  \centering
   \includegraphics[width=\linewidth]{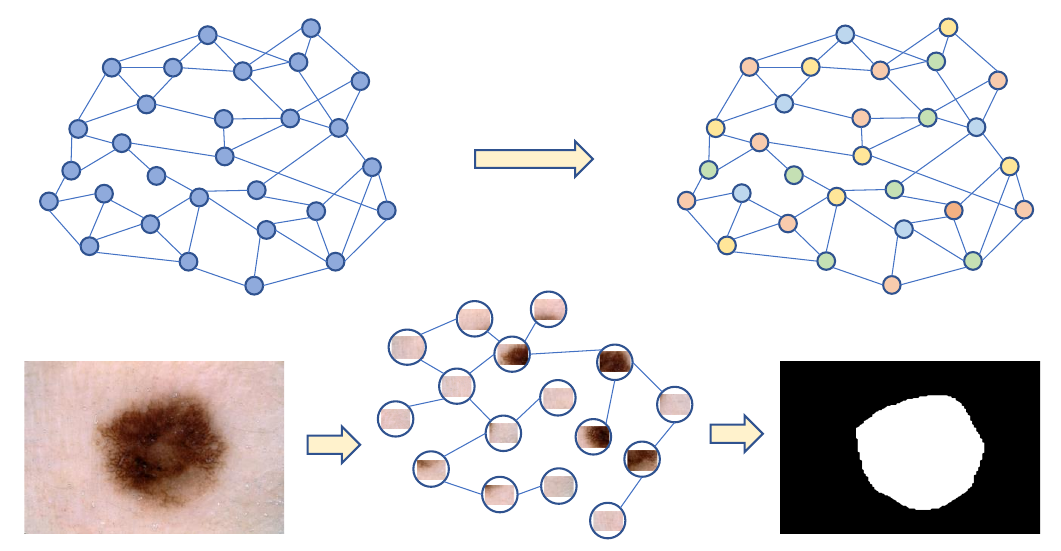}
   \caption{Applications of ChebMixer. Graph node classification and medical image segmentation.}
   \label{fig:application}
\end{figure}  

\section{Introduction}
\label{sec:intro}
Graphs provide an extremely flexible model for approximating the data domains of a large class of problems. For example, computer networks, transportation (road, rail, airplane) networks, or social networks can all be described by weighted graphs, with the vertices corresponding to individual computers, cities, or people, respectively. The remarkable success of deep learning for text \cite{Mikolov2013} and images \cite{Krizhevsky2012} defined on regular domain motivates the study of extensions to irregular, non-Euclidean spaces such as manifold and graph. Therefore, learning representation of irregular data emerges, termed graph neural networks (GNNs) or geometric deep learning \cite{bronstein2017geometric,cao2020comprehensive,bronstein2021geometric, wu2020comprehensive}. With the rapid development of GNNs, they have achieved state-of-the-art performance on almost all tasks among various graph representation learning methods \cite{yao2019graph, wei2023expgcn, wang2019dynamic, han2022vig}. 


GNNs can be roughly divided into two categories: spatial-based and spectral-based. Spatial GNNs \cite{4700287,kipf2017semi,velivckovic2018graph,atwood2016diffusion} imitate the convolution operator in Euclidean space, and the core idea is to aggregate one-hop neighborhood information, which can be quickly implemented via the message-passing scheme. Spatial GNNs possibly learn long-range dependencies by increasing the number of layers but suffer from over-smoothing \cite{huang2020tackling,Oono2020Graph} and over-squashing \cite{topping2022understanding}. Spectral GNNs \cite{defferrard2016convolutional,he2021bernnet,wang2022powerful} are a kind of GNNs that design graph signal filters in the spectral domain of the graph Laplace operator. Theoretically, an arbitrary filter can be represented by choosing a set of basis functions. However, spectral GNNs are often overfitting due to illegal coefficient learning caused by sparse labeling \cite{he2022chebnetii}.

Recently, Transformers emerged as a new architecture and have shown superior performance in a variety of data with an underlying Euclidean or grid-like structure, such as natural language processing \cite{kenton2019bert} and images \cite{dosovitskiy2020image,liu2021swin}. Their great modeling capability motives the generalizing Transformers to non-Euclidean data like graphs. Due to the irregular structure of the graph, the Transformer cannot be directly extended. Graphormer \cite{YingCLZKHSL21}, SAN \cite{kreuzer2021rethinking}, and GPS \cite{rampavsek2022recipe} design powerful positional and structural embeddings to improve their expressive power further. However, graph Transformer generally treats nodes as tokens, which results in quadratic complexity regarding the number of nodes during self-attention computation. Therefore, NAGphormer \cite{chennagphormer} extracts the $K$-hops representation of the node via the adjacent matrix as a preprocessing step and treats each node as a sequence of tokens to directly input the Transformer. However, once input into the Transformer, each node is optimized without considering the relationship between them, which is easy to overfit. Graph MLP Mixer \cite{he2023generalization} extracts graph patches or tokens via graph cluster algorithm and uses the efficient MLP Mixer \cite{tolstikhin2021mlp} to learn token representations from computer vision instead of the self-attention mechanism. However, the clustering algorithm is time-consuming, and the method is unsuitable for graph node and link prediction tasks. More importantly, most graph neural networks are generally validated on graph benchmarks and do not verify performance on other data representations. We believe a unified framework based on graph neural networks is possible because of graphs' flexible and general representation capabilities.

To address these issues, we propose a novel efficient graph neural network and proves its effectiveness in diverse task of different data domain ranging from graph node classification to medical image segmentation. Inspired by NAGphormer \cite{chennagphormer}, we extract multiscale representations of each node and treat each node as a sequence of tokens. One of the influential works to extract multiscale representations of nodes is using spectral graph wavelet transform \cite{hammond2011wavelets} and accelerate computation via spectral filtering based on wavelets filters. For simplicity, we use Chebyshev polynomials as filters to extract the $K$-hop representation of nodes due to their K-localized properties \cite{defferrard2016convolutional}. Instead of costly self-attention computation, we use a more efficient MLP mixer to learn the representations of different-hop neighborhoods based on their semantic correlation. Finally, we aggregate the different-hop neighborhood representations via Chebyshev interpolation to avoid learning illegal coefficients \cite{he2022chebnetii}. Overall, due to the powerful representation capabilities and fast computational properties of MLP Mixer, we can quickly extract more informative node representations benefitting downstream tasks. For validation, we apply our method to tasks from the fields of graph node classification and medical image segmentation and show that it outperforms state-of-the-art approaches. Moreover, it proves that creating a unified architecture via graph representation learning for tasks on different data domains is possible. We summarize our main contributions as follows:
\begin{itemize}
    \item We present a novel graph MLP mixer for graph representation learning, which uses MLP mixer to learn node representations of different-hop neighborhoods, leading to more informative node representation after aggregation. 
    
    \item In addition to the general task of graph node classification, we apply our method to medical image segmentation task. We convert an image to a kNN-graph and learn the image representation based on our proposed graph representation learning approach.  
    
    
    \item The experimental results prove our significant improvements in a variety of scenarios ranging from graph node classification to medical image segmentation. What's more, it proves that creating a unified architecture via graph representation learning for tasks on different data domains is possible.
\end{itemize}


\section{Related Work}
\label{sec:related_work}

\subsection{GNN(Graph Neural Network)}

Graph Neural Networks (GNNs) are a successful tool for graph representation learning, which have achieved state-of-the-art performance on almost all tasks among various graph representation learning methods \cite{yao2019graph, wei2023expgcn, wang2019dynamic, han2022vig}. Generally, GNNs can be roughly divided into two categories: spatial-based and spectral-based.

\textbf{Spatial GNNs} \cite{4700287,kipf2017semi,velivckovic2018graph,atwood2016diffusion} imitate the convolution operator in Euclidean space, and the core idea is to aggregate one-hop neighborhood information, which can be quickly implemented via the message-passing scheme. Scarselli \etal~\cite{4700287} propose the first spatial-based graph convolution method via directly summing the neighborhood information of nodes. And the residual and skip connections are applied to remember the information of each layer. However, it uses a non-standardized adjacency matrix, which leads to hidden node states with different scales. DCNN ~\cite{atwood2016diffusion} considers graph convolution as a diffusion process, where information is transferred from one node to one of its neighboring nodes with a certain probability to balance the information distribution after several rounds. GraphSAGE~\cite{hamilton2017inductive} is a generalized inductive framework that aggregates information by uniformly sampling a fixed-size set of neighbors instead of using the complete set. Message Passing neural network (MPNN)~\cite{gilmer2017neural} utilizes K-step message-passing iterations to propagate information further. To reduce noise and improve performance, some researchers extend the attention operator to the graph by assigning varying weights to neighbors through the attention mechanism \cite{velivckovic2018graph,zhang2018gaan}. Spatial GNNs possibly learn long-range dependencies by increasing the number of layers but suffer from over-smoothing \cite{huang2020tackling,Oono2020Graph} and over-squashing \cite{topping2022understanding}.



\textbf{Spectral GNNs} \cite{defferrard2016convolutional,he2021bernnet,wang2022powerful} are a kind of GNNs that design graph signal filters in the spectral domain of the graph Laplace operator. This method is first proposed by Bruna \etal~\cite{bruna2014spectral} but suffers from costly graph Fourier transform. He \etal~\cite{he2021bernnet} categorizes spectral GNNs based on the type of filtering operation used into two categories. The first category comprises spectral domain GNNs with fixed filters, such as APPNP~\cite{klicpera2019predict}, which uses personalized PageRank to construct the filter function. The second category includes spectral domain GNNs with learnable filters. ChebNet~\cite{defferrard2016convolutional} utilizes diagonal matrices based on Chebyshev-polynomial eigenvalue expansions to approximate graph convolution. GPRGNN~\cite{chien2021adaptive} employs direct gradient descent on polynomial coefficients. Additionally, some works use other polynomial bases~\cite{he2021bernnet, bianchi2021graph,  zhu2021interpreting, wang2022powerful}, such as BernNet~\cite{he2021bernnet} which uses the Bernstein polynomial basis, ARMA~\cite{bianchi2021graph} and GNN-LF/HF\cite{zhu2021interpreting} which utilize rational functions. Among the mentioned methods, GPRGNN and ChebNet are considered more expressive because the former can express all polynomial filters, and the latter uses Chebyshev polynomials that can form a complete set of bases in polynomial space. ChebNet often suffers from overfitting issues in experimental settings despite being theoretically more expressive. To alleviate the overfitting problem in ChebNet, Kipf \etal propose GCN~\cite{kipf2017semi}, simplifying ChebNet by using filters that run on 1-hop neighborhoods of the graph. He \etal~\cite{he2022chebnetii} theoretically demonstrates that the overfitting issue of ChebNet is caused by the illegal parameters it learns. To address this problem, they propose ChebNetII, which enhances the approximation of the original Chebyshev polynomials using Chebyshev interpolation.


\subsection{Graph Transformer and MLP-Mixer}
\textbf{Graph Transformers} emerges as a new architecture and has shown superior performance on various graph mining tasks recently. The core idea of graph Transformer is to learn node representations by integrating graph structures into Transformer structures. Recognizing the similarity between the attention weights and the weighted neighbor matrix of a fully connected graph, Dwivedi \etal~\cite{graphtransformer} combine Transformer with GNN. Several studies have explored strategies to replace the traditional position encoding in the Transformer with powerful position and structure encodings based on relevant graph information. For instance, SAN~\cite{kreuzer2021rethinking}, LSPE~\cite{dwivedi2022graph}, NAGphormer~\cite{chennagphormer} use the Laplacian operator or random walk operator to learn the position encoding. Graphormer~\cite{YingCLZKHSL21} develops the centrality encoding according to the degree centrality, the spatial encoding and the edge encoding according to the shortest path distance. In general, most graph Transformers can solve the issue of over-squeezing and limited long-range dependency in GNNs by using a non-local self-attention mechanism. However, they also increase the complexity from $O(E)$ to $O(N^2)$, resulting in computational bottlenecks.



Due to the high computational cost of the Transformer, this work has recently been challenged by the more time-efficient novel model called MLP-Mixer~\cite{tolstikhin2021mlp}. Instead of convolutional or attentional mechanisms, MLP-Mixer uses multi-layer perceptron unaffected by over-squeezing or limited long-range dependency. MLP-Mixer is initially used for image-related tasks such as image classification~\cite{tolstikhin2021mlp, touvron2022resmlp, zhang2022multi}, image segmentation~\cite{lai2022axial} and other tasks. However, a generalization of MLP mixer to graph is challenging given the irregular and variable nature of graphs. He \etal \cite{he2023generalization} successfully generalize MLP mixer to graph via graph clustering algorithm to extract graph patch or token. However, the clustering algorithm is time-consuming, and the method is unsuitable for graph node and link prediction tasks.



\begin{figure*}
    \centering
    \includegraphics[width=0.9\linewidth]{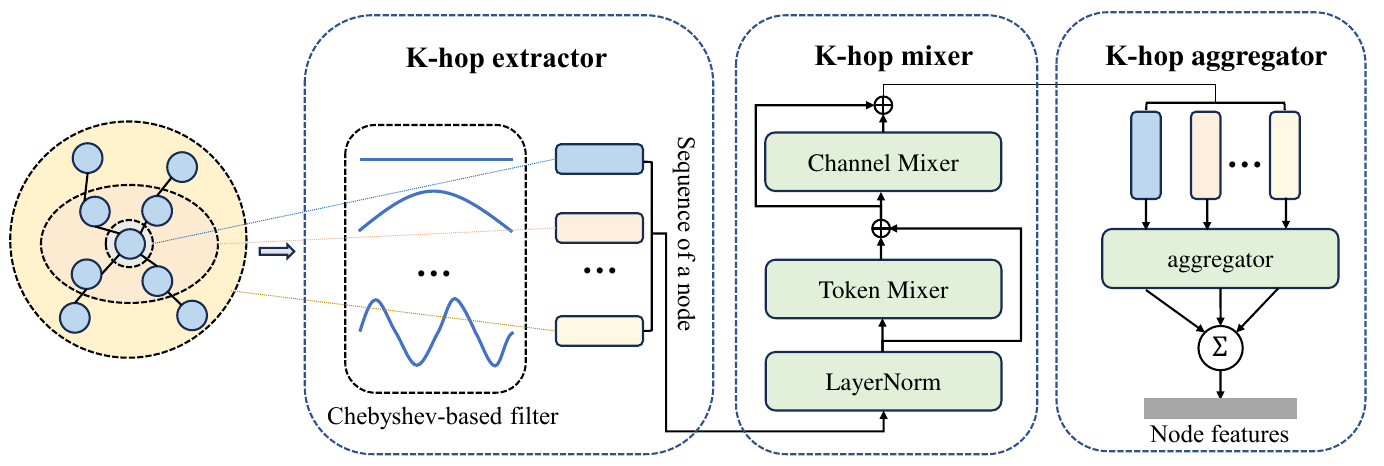}
    \caption{\textbf{The proposed  ChebMixer.} ChebMixer first uses a novel neighborhood extraction module, $K$-hop extractor, to generate multiscale or multi-hop representations of graph nodes and treat them as a sequence of tokens. ChebMixer then refines the multi-hop representations of graph node with an effective MLP Mixer and develops a novel aggregator to aggregate the multiscale representations of the nodes.} 
  \label{fig:ChebMixer}
\end{figure*}

\section{ChebMixer}
In this section, we first introduce some essential knowledge related to graphs and spectral filtering. After that, we introduce the proposed ChebMixer in detail.

\subsection{Preliminary}
\noindent\textbf{Graph}. A graph is composed of a finite non-empty set of vertices and a set of edges, which can be represented as $G = (V, E)$, where $V$ is a finite set of $|V|=N$ vertices, $E$ is a set of edges and $A \in \mathbb{R}^{N \times N}$ is a weighted adjacency matrix encoding the connection weight between two nodes. In graph machine learning and graph neural networks, a node feature can be regarded as a matrix $X \in \mathbb{R}^{N \times d}$. Here, $d$ is the feature dimension of each node. Expressly, $A_{i,j}$ is set to 1 or the edge's weight if the edge exists from node $i$ to node $j$; otherwise, it is set to 0.

\textbf{Spectral convolution} is a graph convolution method based on the spectral graph theory \cite{chung1997spectral}. It involves converting the graph signal into the spectral domain for processing through a spectral filter, which is then converted back into the spatial domain. Spectral convolution is based on the graph Laplacian Laplacian $L$, which combinatorial definition is $L = D - A$ where $D \in \mathbb{R}^{N \times N}$ is the diagonal degree matrix with $D_{ii}=\sum_j{A_{ij}}$, and symmetric normalized definitions is $L = I_N - D^{-1/2}AD^{1/2}$ where $I_N$ is the identity matrix. As $L$ is a real symmetric positive semidefinite matrix, it can be decomposed as $L = U \Lambda U^T $, where $U$ is the eigenvector matrix of the Laplacian matrix, $\Lambda$ is the diagonal matrix with the eigenvalues of the Laplacian matrix as its diagonal elements. After that, the spectral domain convolution filters the eigenvalues by designing different filters. Given spectral filter $h(\lambda)$, the spectral convolution operator can be defined as
\begin{equation}
    Y=Uh(\Lambda)U^TX 
  \label{eq:spec_filter}
\end{equation}
where $y$ denotes the filtering results of $x$. Many methods to learn the optimal spectral filters via polynomials expansions  $h(\lambda) \approx \sum^K_{k=0}\theta_k\lambda^k$ , where coefficients $\theta_k$ of expansions are trainable. So we have 
\begin{equation}
    Y = Uh(\Lambda)U^TX \approx U\sum^K_{k=0}\theta_k\Lambda^kU^TX  \approx \sum^{K}_{k=0}\theta_k L^kX 
  \label{eq:polyfilter}
\end{equation} 

ChebNet~\cite{defferrard2016convolutional} is an outstanding contribution to the polynomial spectral convolution field, using Chebyshev polynomials to implement filtering, as shown below:
\begin{equation}
  Y \approx \sum^{K}_{k=0}\theta_k T_k(\hat{L})X
  \label{eq:chev_filter}
\end{equation}
where $T_k(\hat{L}) \in \mathbb{R}^{N \times N}$ is the Chebyshev polynomial of order $k$ evaluated at $\hat{L}$. $\hat{L}$ denotes the scaled Laplacian calculated by $\hat{L} = 2L/\lambda_{max} - I_N$. The Chebyshev polynomials $T_k(x)$ of order k can be recursively calculated by $T_k(x) = 2xT_{k-1}(x) - T_{k-2}(x)$ with $T_0(x) = 1$ and $T_1(x) = x$. Due to the K-localized properties of $T_k(\hat{L})$, we can reinterpret the Eq.(\ref{eq:chev_filter}) as two steps: (1) extract $K$-hop neighborhood representations;(2) aggregate the different-hop neighborhoods representations. Our core idea is to use effective MLP mixer to enhance the $K$-hop neighborhood representations for more informative node features. 


\begin{figure*}
    \centering
    \includegraphics[width=\linewidth]{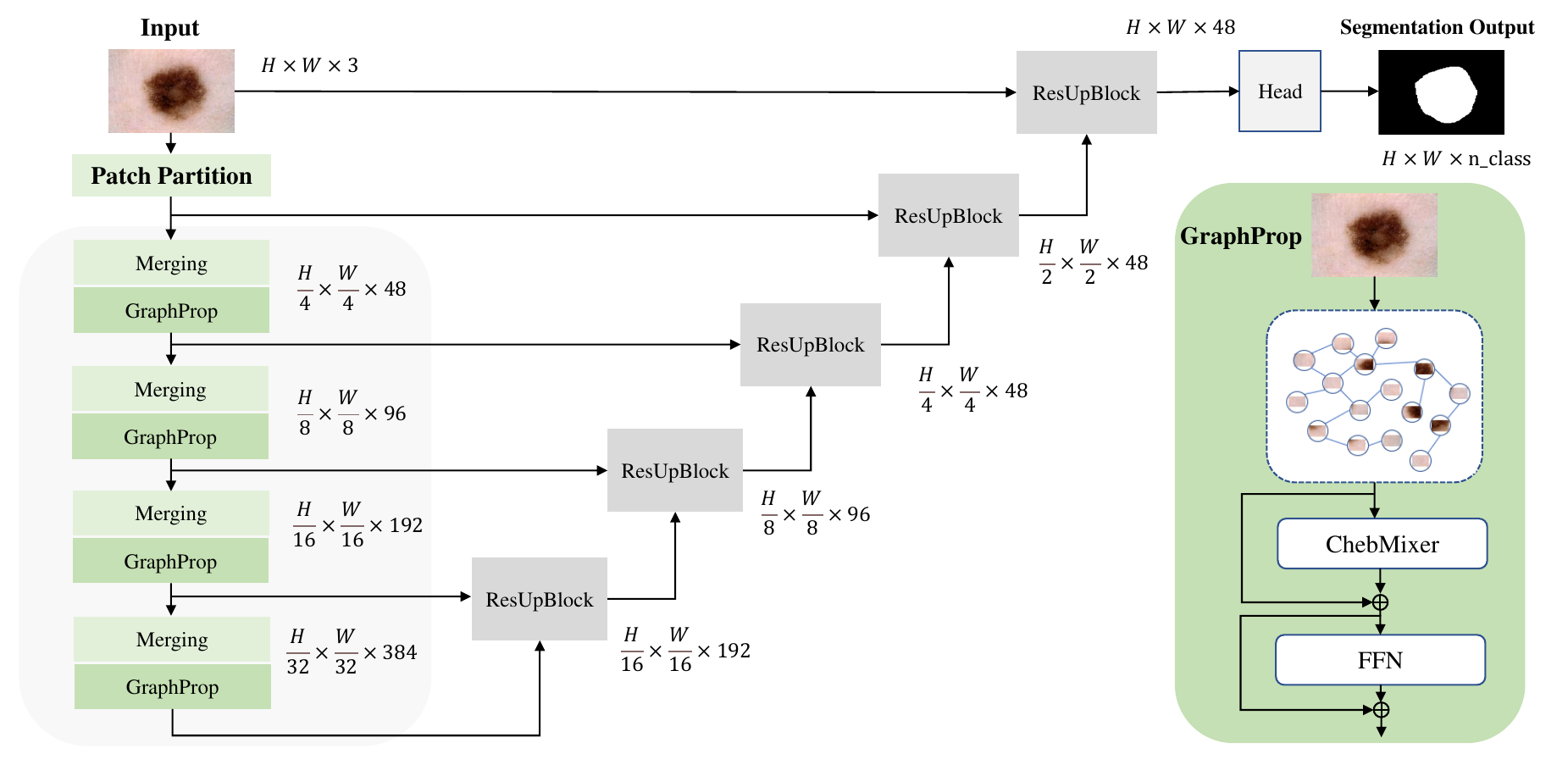}
     \caption{\textbf{Architecture for medical image segmentation}, which is composed of encoder, decoder, and skip connections. The encoder contains two modules: patch merging layer and GraphProp. The former performs downsampling, and the latter is constructed based on ChebMixer for feature extraction. The encoded feature representations in each encoder layer are fed into the corresponding CNN-decoder(\emph{i.e.}, the ResUpBlock in the figure, which has the function of up-sampling) via skip connections. The final output of the decoder goes through a head to obtain the segmentation result.} 
     \label{fig:segmentation}
\end{figure*}

\subsection{Methods}
\label{sec:methods}
The overview of our method is introduced in \cref{fig:ChebMixer}. It includes three parts: extracting the $K$-hop representations of nodes via fast Chebyshev polynomials spectral filtering, mixing the $K$-hop representations of nodes with effective MLP  mixer and aggregating $K$-hop representations via Chebyshev interpolation. 

\noindent\textbf{$K$-hop extractor}. We design a module for $K$-hop neighborhood computation based on Chebyshev polynomials. Specifically, for input adjacency matrix $A \in \mathbb{R}^{N \times N}$ and node feature matrix $X \in \mathbb{R}^{N \times d}$, we first compute the scaled normalized Laplacian matrix $\hat{L}$. Due to the recurrence relation of Chebyshev polynomials, we can compute the $K$-hop neighborhood feature embedding of all nodes as \cref{alg:k_extractor}. The final output is $X_G = (X^0, X^1,..., X^k, ..., X^K)$, where $K$ is a hyperparameter, $X^k \in \mathbb{R}^{N \times d}$ denotes the $k$-th hop neighborhood information, and $X_G \in \mathbb{R}^{N \times (K + 1) \times d}$ denotes the sequence of $K$-hop neighborhood features of all nodes.
\begin{algorithm}
\caption{$K$-hop extractor based on Chebyshev.}
\label{alg:k_extractor}
\begin{algorithmic}[1]
\REQUIRE Scaled Normalized Laplacian Matrix $\hat{L}$; Feature matrix $X\in \mathbb{R}^{N \times d}$; Chebyshev polynomials order $K$
\ENSURE The $K$-hop neighborhood representation of all nodes $X_G \in \mathbb{R}^{N \times ((K+1)) \times d}$
\STATE $X_G[:,0,:] = X,  X_G[:,1,:] = \hat{L}X$
\FOR{$i=2$ to $K$}
    \STATE $X_G[:,i,:] = 2\hat{L}X_G[:,i-1,:] - X_G[,i-2,:]$
\ENDFOR
\RETURN $X_G \in \mathbb{R}^{N \times (K+1) \times d}$
\end{algorithmic}
\end{algorithm} 

\noindent\textbf{$K$-hop mixer}. Having obtained the information sequence $X_G \in \mathbb{R}^{N \times(K+1)\times d}$ for the $K$-hop neighborhood, we treat each node as a sequence of tokens and follow the approach of Tolstikhin \etal~\cite{tolstikhin2021mlp}, alternating channel and $K$-hop mixing steps using a simple mixer layer. Token mixing is used for the $K$-hop dimension, while channel mixing is used for the channel dimension. The mixer layer can be expressed as
\begin{equation}
    \begin{aligned}
        X_G &= X_G + (W_2 \sigma (W_1 LayerNorm(X_G)))   \\
        X_G &= X_G + (W_4 \sigma (W_3 LayerNorm(X_G)^T))^T 
    \end{aligned}
  \label{eq:mixing}
\end{equation}
where LayerNorm is layer normalization~\cite{ba2016layer}, $\sigma$ is a nonlinear activation function~\cite{hendrycks2016gaussian}, and matrix $W_1 \in  \mathbb{R} ^{d_s \times (K+1)}$, $W_2 \in  \mathbb{R} ^{(K+1)\times d_s}$, $W_3 \in  \mathbb{R} ^{d_c \times d}$, $W_4 \in  \mathbb{R} ^{d \times d_c}$, both $d_s$ and $d_c$ are hyperparameters.

\noindent\textbf{$K$-hop aggregator}. Finally, we aggregate the sequence of $K$-hop neighborhood features $X_G$ to produce more informative node features. To alleviate the illegal coefficients learning, we utilize method in ChebNetII~\cite{he2022chebnetii}, which reparameterizing the learning coefficients via Chebyshev interpolation. In contrast to ChebNetII, using the same parameters for all channels, we assert that different channels represent different information and should have different aggregation weights. Hence, we propose an aggregator with a set of learnable weights $W \in \mathbb{R} ^ {(K+1)\times d}$. We restrict $W$ according to \cref{eq:cheb_interpolation} and aggregate $K$-hop neighborhood information sequences based on \cref{eq:aggregator}.

\begin{equation}
    W^k = \frac{2}{K + 1}\sum_{j=0}^{K} \gamma_{j}T_k(x_j)
    \label{eq:cheb_interpolation}
\end{equation}

\begin{equation}
    X_{agg} = \sum_{k=0}^{K}W^k \times X_G^k 
    \label{eq:aggregator}
\end{equation}
where $W^k$ denotes the vector of learnable weights $W$ at index $k$, $X_G^k $ denotes the matrix of the sequence of feature matrices at index $k$, $\gamma_j \in \mathbb{R}^d$ is the learnable parameter, and $x_j = cos((j + 1/2)\pi/(K+1))$ is a Chebyshev node of $T_{K+1}$. The detailed implementation is drawn in \cref{alg:k_agg}.

\begin{algorithm}
\caption{$K$-hop aggregator}
\label{alg:k_agg}
\begin{algorithmic}[1]
\REQUIRE Learnable weight $W \in \mathbb{R} ^ {(K+1) \times d}$; Sequence of feature matrices $X_G \in R^{N \times ((K+1)) \times d}$
\ENSURE Aggregated node feature matrix $X_{agg} \in R^{N \times d}$
\STATE Clone $W$ to $\gamma$ ; Initialize $X_{agg}$ with 0
\FOR{$k=0$ to $K$}
    \STATE $W[k] = \gamma[0] \times T_k(x_0)$
    \FOR{$j = 1$ to $K$}
        \STATE $W[k] = W[k] + \gamma[j] \times T_k(x_j)$
    \ENDFOR
    \STATE $W[k] = 2W[k] / (K + 1)$
\ENDFOR
\FOR{$k=0$ to $K$}
    \STATE $X_{agg} = X_{agg} + X_G[:,k,:] \times W[k]$
\ENDFOR
\RETURN Aggregated node feature matrix $X_{agg} \in \mathbb{R}^{N \times d}$
\end{algorithmic}
\end{algorithm}

\begin{table*}[h!t]
  \centering
  \begin{tabular}[t]{lccccccc}
    \toprule
    Method & Cora & Citeseer & Pubmed & Computers & Photo & CoauthorCS & ogbn-arxiv\\
    \midrule
    MLP       
    & 72.91 ± 0.16& 76.18 ± 0.70&84.08 ± 0.11&74.20 ± 0.85&74.89 ± 0.42& 94.51 ± 0.15& 51.58 ± 0.14           \\
    GCN~\cite{kipf2017semi}       
    &88.61 ± 0.20 &79.38 ± 0.33&86.34 ± 0.08&79.55 ± 0.11& 88.47 ± 0.20& 94.61 ± 0.03& 66.19 ± 0.57           \\
    GAT~\cite{velivckovic2018graph}       
    &86.35 ± 0.40&79.48 ± 0.66&85.98 ± 0.20&81.56 ± 0.63& 86.96 ± 0.37& 93.38 ± 0.18& 64.86 ± 0.34           \\
    APPNP~\cite{klicpera2019predict}    
    &88.65 ± 0.20&79.33 ± 0.84&85.79 ± 0.21&72.29 ± 0.20&84.31 ± 1.72&94.07 ± 0.10&65.98 ± 0.29            \\
    GPRGNN~\cite{chien2021adaptive}    
    &89.29 ± 0.36&80.33 ± 0.09&87.94 ± 0.28&89.08 ± 0.96&94.33 ± 0.36&95.13 ± 0.06&69.86 ± 0.70           \\
    ChebNet~\cite{defferrard2016convolutional}   
    &84.94 ± 0.32&78.16 ± 0.42&88.25 ± 0.11&90.60 ± 0.13&93.88 ± 0.14&94.49 ± 0.11&70.79 ± 0.20            \\
    ChebNetII~\cite{he2022chebnetii} 
    &\textbf{89.69 ± 0.64}&80.94 ± 0.37&88.93 ± 0.29&81.76 ± 2.27&89.65 ± 0.69&94.53 ± 0.28&72.32 ± 0.23\\
    NAGphormer~\cite{chennagphormer} 
    &87.68 ± 0.52&77.12 ± 0.80&89.02 ± 0.19&91.22 ± 0.14&95.49 ± 0.11&\textbf{95.75 ± 0.09}&71.01 ± 0.13            \\
    Ours   
    &{\color{blue}89.46 ± 0.24}&\textbf{81.04 ± 0.28}&\textbf{89.22 ± 0.35}&\textbf{92.95 ± 0.20}&\textbf{96.19 ± 0.17}&{\color{blue}95.53 ± 0.13}&\textbf{73.28 ± 0.06}            \\ 
    \bottomrule
  \end{tabular}
  \caption{Comparison of all models in terms of mean accuracy ± stdev (\%) on graph node classification. The best results appear in bold. Blue indicates that the results of our model are comparable to the best results.}
  \label{tab:node_classification}
\end{table*}

\subsection{Network Architecture Details}
The ChebMixer module we developed is illustrated in \cref{fig:ChebMixer}. To evaluate the effectiveness of our approach, we construct different network architectures and experiment on both graph node classification and medical image segmentation tasks. For graph node classification, we begin by feeding the node feature matrix $X$ into a linear projection layer to map the node feature dimensions to $d$, followed by sequentially updating the features in the three modules described in \cref{sec:methods}. The resulting output is then passed through a classification header for classification. 

For medical image segmentation, we construct a UNet-like architecture using the architectures in ViG~\cite{han2022vig} and SwinUNETR~\cite{hatamizadeh2021swin},  as shown in \cref{fig:segmentation}. It comprises encoder, decoder, and skip connections. The encoder has four layers, each with a GraphProp module based on ChebMixer. This module transforms the image into a graph and extracts features. Specifically, for an image with size of $H \times W \times 3$, we first divide it into $N$ patches. By converting each patch into a feature vectors $x_i \in \mathbb{R} ^ d$, we obtain  a set of feature matrices $X = (x_1, x_2,... ,x_N)$. Each patch can be considered as a node, resulting in a set of node sets $V$. After that, for each node $v_i \in V$ (\emph{i.e.}, patch), we construct $K$-nearest neighbors graph in feature space. By considering the image as a graph, we can extract the node representation using the proposed ChebMixer module. To maintain the hierarchical structure of the encoder,  a patch merging layer is used before each stage for performing downsampling by $2 \times$ and doubling the feature dimensions from their original size. The feature representations extracted by the encoder are used in the decoder via skip connections at each layer. Each layer of the decoder contains a module named ResUpBlock, which performs upsampling using convolution. This process reshapes the feature maps of adjacent dimensions into feature maps with a resolution of $2 \times$ and reduces the feature dimension to half of the original dimensions accordingly.

\section{Results}
This section introduces our experiments and results in detail. Our models are implemented by PyTorch, and all the experiments are carried out on a machine with an NVIDIA RTX3090 GPU ( 24GB memory ), Intel Xeon CPU ( 2.1 GHz ) with 16 cores, and 64 GB of RAM.
\subsection{Graph Node Classification}
\label{sec:gnc}
\noindent\textbf{Datasets}.
 We evaluate ChebMixer on seven widely used real-world datasets, which include six small-scale datasets: Cora, Citeseer, Pubmed, Computers, Photo, CoauthorCS, and a large-scale reference dataset: ogbn-arxiv. We apply 60\%/20\%/20\% train/val/test random splits for small-scale datasets. For ogbn-arxiv, we use the partition method in OGB~\cite{hu2020ogb}.

\noindent\textbf{Baselines and settings}. We compare ChebMixer with 8 advanced baselines, including GCN~\cite{kipf2017semi}, GAT~\cite{velivckovic2018graph}, APPNP~\cite{klicpera2019predict}, GPRGNN~\cite{chien2021adaptive},  MLP, ChebNet~\cite{defferrard2016convolutional}, ChebNetII~\cite{he2022chebnetii}, NAGphormer~\cite{chennagphormer}. For all models, We employ the AdamW~\cite{loshchilov2018decoupled} optimizer with an early stopping of 50 and a maximum of 2000 epochs to train.

\noindent\textbf{Results}. We rigorously conducted five trials for each model, ensuring that each trial's random seed is different to minimize potential bias. The results are shown in \cref{tab:node_classification}. From the experimental results, we can observe that ChebMixer outperforms the baseline on the above dataset, proving our proposed ChebMixer's superiority.
\begin{table}
    \centering
    \begin{threeparttable}
    \begin{tabular}[t]{lccr}
     \toprule
     Model & Dice  $\uparrow$ & IOU $\uparrow$ & \makecell{Params \\ (M)} \\
    \midrule
    UTNet~\cite{gao2021utnet}                & 89.04 & 80.88 & 10.02 \\
    BAT~\cite{wang2021boundary}              & 90.03 & 82.41 & 46.73 \\
    TransFuse~\cite{zhang2021transfuse}      & 90.31 & 82.87 & 26.27 \\
    SwinUnet~\cite{cao2022swin}              & 89.55 & 81.75 & 41.39\\
    SwinUNETR~\cite{hatamizadeh2021swin}     & 90.26 & 82.77 & 25.13\\
    HiFormer~\cite{heidari2023hiformer}     & 90.46  & 83.07 & 23.37\\ \hline
    Base\tnote{*}                           & 90.42 & 83.04 & 5.23  \\
    Ours                          & \textbf{91.68} & \textbf{84.57} & 7.60 \\
    \bottomrule
    \end{tabular}
    \begin{tablenotes}
        \footnotesize
        \item[*] Base is a UNet-like architecture based on ViG~\cite{han2022vig}.

      \end{tablenotes}
    \end{threeparttable}
    \caption{Comparison of segmentation results between our model, Base, and SOTA methods. All models are validated by five-fold cross-validation and reported mean values(\%). We report the models' parameter count in millions (M).}
    \label{tab:segmentation}
\end{table}


\subsection{Medical Image Segmentation}
\noindent\textbf{Datasets and evaluation metrics}.
We conduct experiments on the skin disease segmentation dataset ISIC2018~\cite{codella2019skin}, which contains 2594 images and corresponding segmentation masks. We resize the images to 256 × 256 and enhance them with random flipping, rotation, Gaussian noise, contrast, and brightness variations. 

\noindent\textbf{Implementation details}. We train the model by combining binary cross entropy (BCE) and dice loss. The loss   $\mathcal{L}$ between $\hat{y}$ and the target y can be formulated as :
\begin{equation}
       \mathcal{L}= B C E(\hat{y}, y)+\operatorname{Dice}(\hat{y}, y)
  \label{eq:se_loss}
\end{equation}
We train models for 200 epochs with AdamW~\cite{loshchilov2018decoupled} optimizer. The learning rate is initialized to $ 1 \times 10 ^ { -4 } $ and changed by a linear decay scheduler with a step size of 50 and decay factor $ \gamma = 0.5 $. We evaluate models by Dice and IOU metrics. 

\begin{figure}[t]
  \centering
   \includegraphics[width=\linewidth]{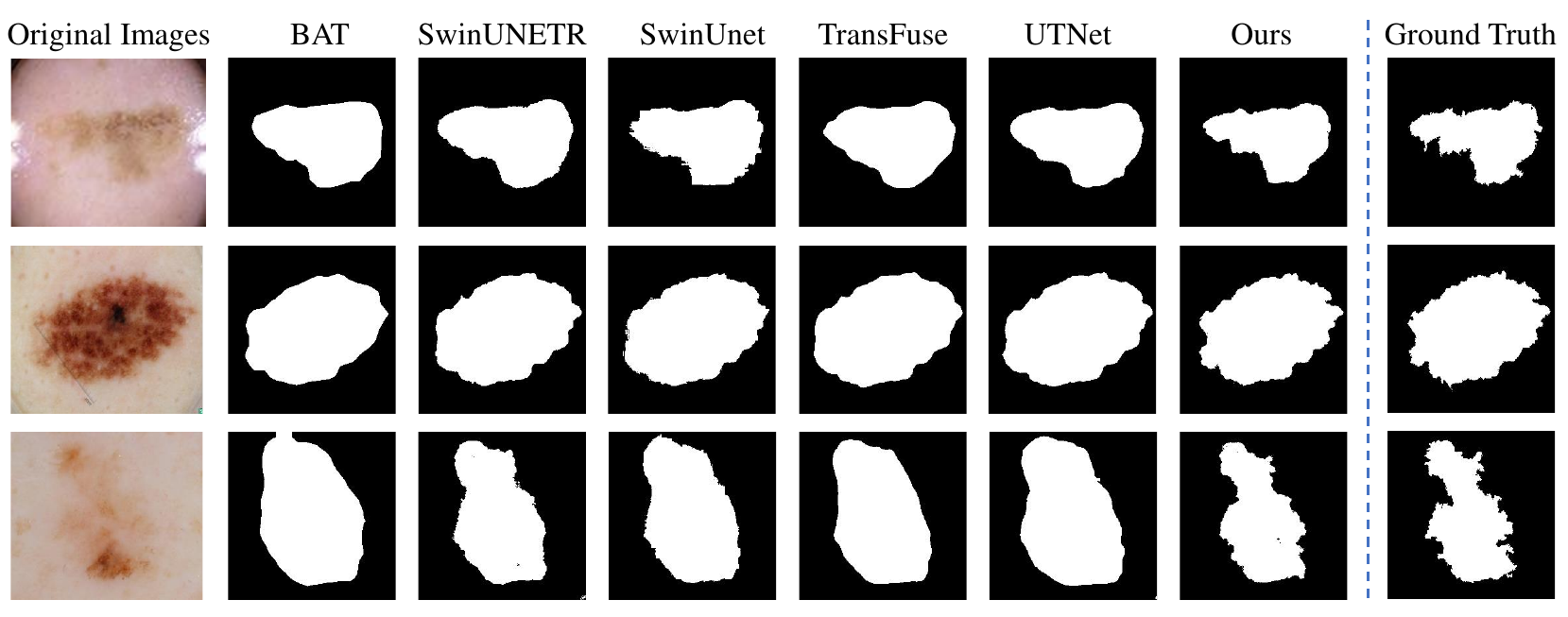}
   \caption{Visual result comparison of our model and SOTA.}
   \label{fig:qc_segmentation}
\end{figure}

\noindent\textbf{Comparisons with the state-of-the-arts}.
We compare our model with BASE and six other SOTA models. BASE is a UNet-like architecture based on ViG~\cite{han2022vig}, which is consistent with our model. The difference is that the graph convolution module of BASE adopts EdgeConv~\cite{wang2019dynamic}. The SOTA includes UTNet~\cite{gao2021utnet}, BAT~\cite{wang2021boundary}, TransFuse~\cite{zhang2021transfuse}, SwinUnet~\cite{cao2022swin}, SwinUNETR~\cite{hatamizadeh2021swin}, HiFormer~\cite{heidari2023hiformer}. All models are validated by five-fold cross-validation and reported mean values. The experimental results are recorded in \cref{tab:segmentation}, which show that our model can achieve better results with fewer parameters. We also show a visual comparison of the skin lesion segmentation results in \cref{fig:qc_segmentation}, demonstrating that our proposed method can capture finer structures and generate more accurate contours.

\subsection{Ablation Studies}
To evaluate the effectiveness of our model, we conducted a series of experiments on graph node classification. Specifically, we conduct ablation experiments on each module of ChebMixer and explore the influence of polynomial order $K$ on the resultant outcomes. These experiments aim to provide an in-depth evaluation of the proposed model architecture's performance and gain insights into its underlying mechanisms.

\noindent\textbf{$K$-hop extractor.} This module is used to extract $K$-hop neighborhood information in the model. To evaluate the efficacy of this module, we compare it with the hop2Token module in NAGphormer~\cite{chennagphormer}. The remaining modules and experimental settings are consistent with \cref{sec:gnc}, and the experimental results are shown in \cref{table:abs_k_hop_extractor}. Our analysis of the experimental results indicates that our module has demonstrated varying degrees of improvement across all datasets except for Citeseer and CoauthorCS. In addition, our module has exhibited a lower standard deviation on all datasets, which suggests greater stability.

\begin{table}[h!t]
\begin{tabular}{lccc}
    \toprule
           &  Hop2Token   &      Ours    & $\Delta$ \\ 
    \midrule
Cora       & 89.01 ± 0.64 & 89.46 ± 0.24 &  + 0.45      \\
Citeseer   & 81.32 ± 2.07 & 81.04 ± 0.28 &  - 0.28    \\
Pubmed     & 89.03 ± 0.42 & 89.22 ± 0.35 &  + 0.19     \\
Computers  & 91.64 ± 0.17 & 92.95 ± 0.20 &  \textbf{+ 1.31}      \\
Photo      & 95.82 ± 0.21 & 96.19 ± 0.17 &  + 0.37      \\
CoauthorCS & 95.59 ± 0.20 & 95.53 ± 0.13 &  - 0.06     \\
ogbn-arxiv & 72.91 ± 0.11 & 73.28 ± 0.06 &  + 0.37    \\
    \bottomrule
\end{tabular}
\caption{Ablation experiment on $K$-hop extractor. $\Delta$ denotes the change in performance, where ``+" denotes an increase in performance and ``-" denotes a decrease in performance.}
\label{table:abs_k_hop_extractor}
\end{table}

\noindent\textbf{$K$-hop mixer.} 
We conduct a comparative analysis of our proposed model with and without the $K$-hop mixer module while keeping other modules and experimental settings consistent with section \cref{sec:gnc}. As depicted in \cref{table:abs_k_hop_mixer}, the experimental results reveal a substantial improvement in the model's performance when the $K$-hop mixer module is incorporated. This finding underscores the significance of updating features in the $K$-hop neighborhood, which is essential for obtaining comprehensive information and improving the model's performance.


\begin{table}[h!t]
\begin{tabular}{lccc}
    \toprule  
          &  \ding{56}   &      \ding{52}    & $\Delta$ \\ 
    \midrule
Cora       & 85.59 ± 0.88 & 89.46 ± 0.24 &  \textbf{+ 3.87}      \\
Citeseer   & 81.17 ± 0.61 & 81.04 ± 0.28 &  - 0.13    \\
Pubmed     & 88.77 ± 0.10 & 89.22 ± 0.35 &  + 0.45     \\
Computers  & 89.28 ± 0.27 & 92.95 ± 0.20 &  \textbf{+ 3.67}      \\
Photo      & 94.71 ± 0.41 & 96.19 ± 0.17 &  + 1.48      \\
CoauthorCS & 94.47 ± 0.28 & 95.53 ± 0.13 &  + 1.06     \\
ogbn-arxiv & 72.53 ± 0.17 & 73.28 ± 0.06 &  + 0.75    \\
    \bottomrule
\end{tabular}
\caption{Ablation experiment on $K$-hop mixer. \ding{52} represents experiments with the mlp mixer, while \ding{56} represents experiments without the mlp mixer.}
\label{table:abs_k_hop_mixer}
\end{table}

\begin{figure*}[h!t]
  \centering
   \includegraphics[width=\linewidth]{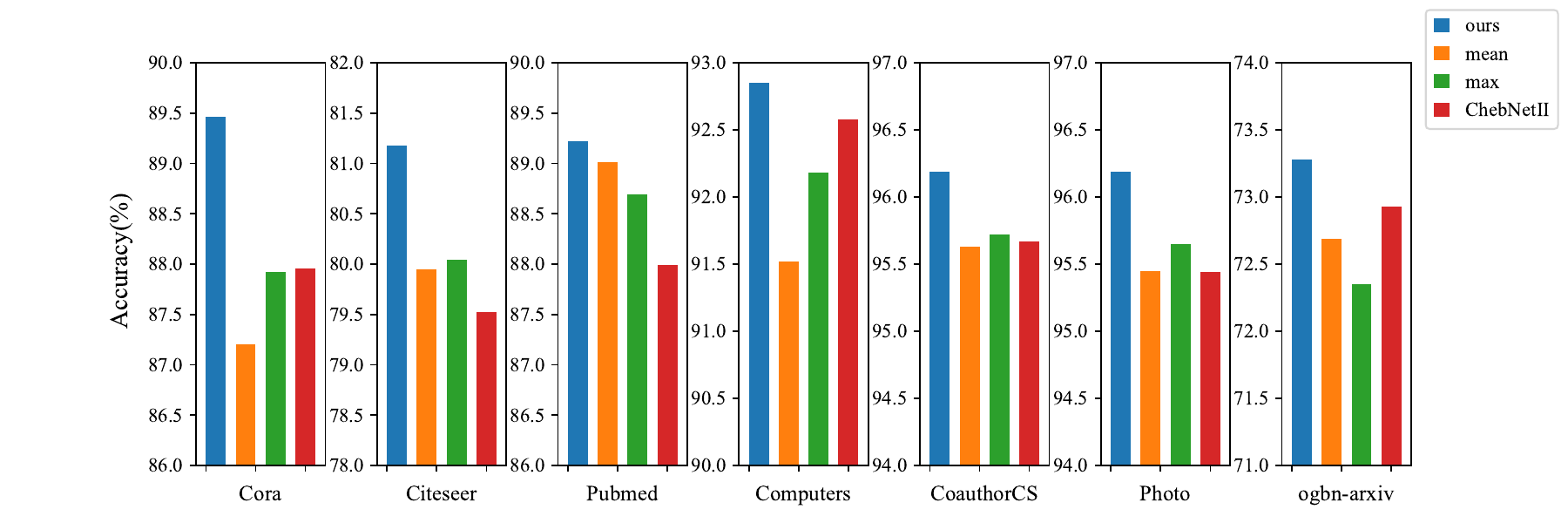}
   \caption{Ablation experiment on $K$-hop aggregator. The "ChebNetII" represents the aggregation module in ChebNetII~\cite{he2022chebnetii}.}
   \label{fig:abs_aggregrate}
\end{figure*}

\begin{figure}[h!t]
  \centering
   \includegraphics[width=\linewidth]{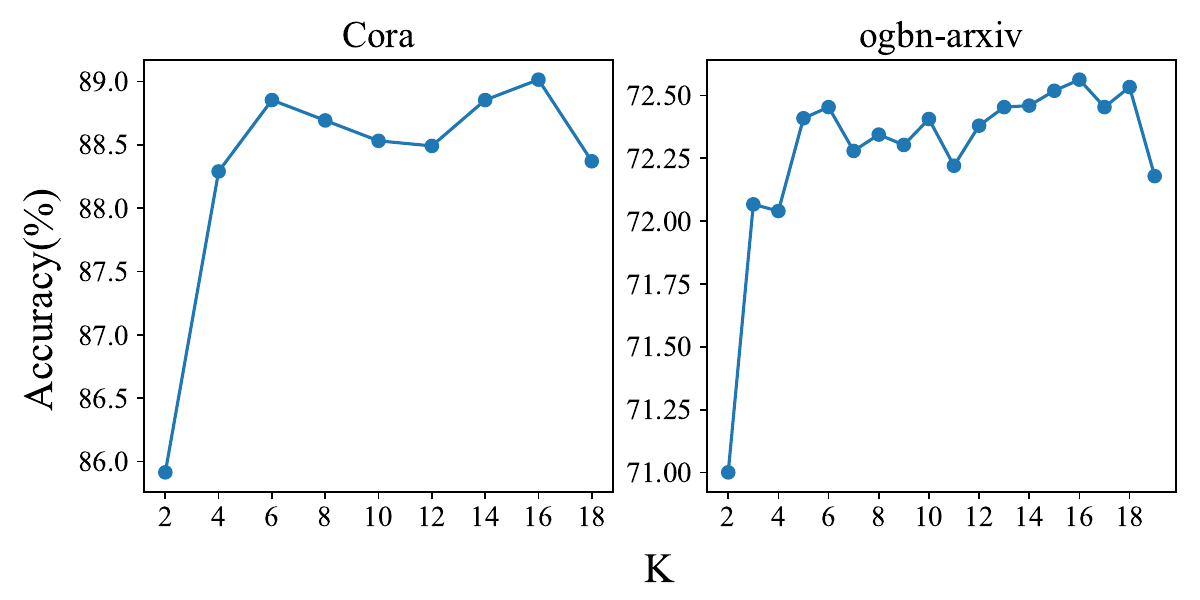}
   \caption{The influence of order of polynomials  $K$.}
   \label{fig:k}
\end{figure}

\noindent\textbf{$K$-hop aggregator.} To verify the effectiveness of the aggregation module, we compare the summation, averaging, maximization, and aggregation methods in ChebNetII~\cite{he2022chebnetii}. The experimental results, as depicted in \cref{fig:abs_aggregrate}, demonstrate the effectiveness of our designed aggregator and highlight the advantages of learning distinct weights on various channels for more informative aggregation.

\noindent\textbf{The influence of order of polynomials  $K$.} 
We experiment with different values of $K$ on various graph node classification datasets. As depicted in \cref{fig:k}, the experimental results show that when $K$ is small (less than 6), the increase of $K$ will significantly improve the model's performance. As $K$ is further increased, the model's performance exhibits a jittery but gradually rising trend. 


\subsection{Runtime}
\begin{figure}[t]
  \centering
   \includegraphics[width=\linewidth]{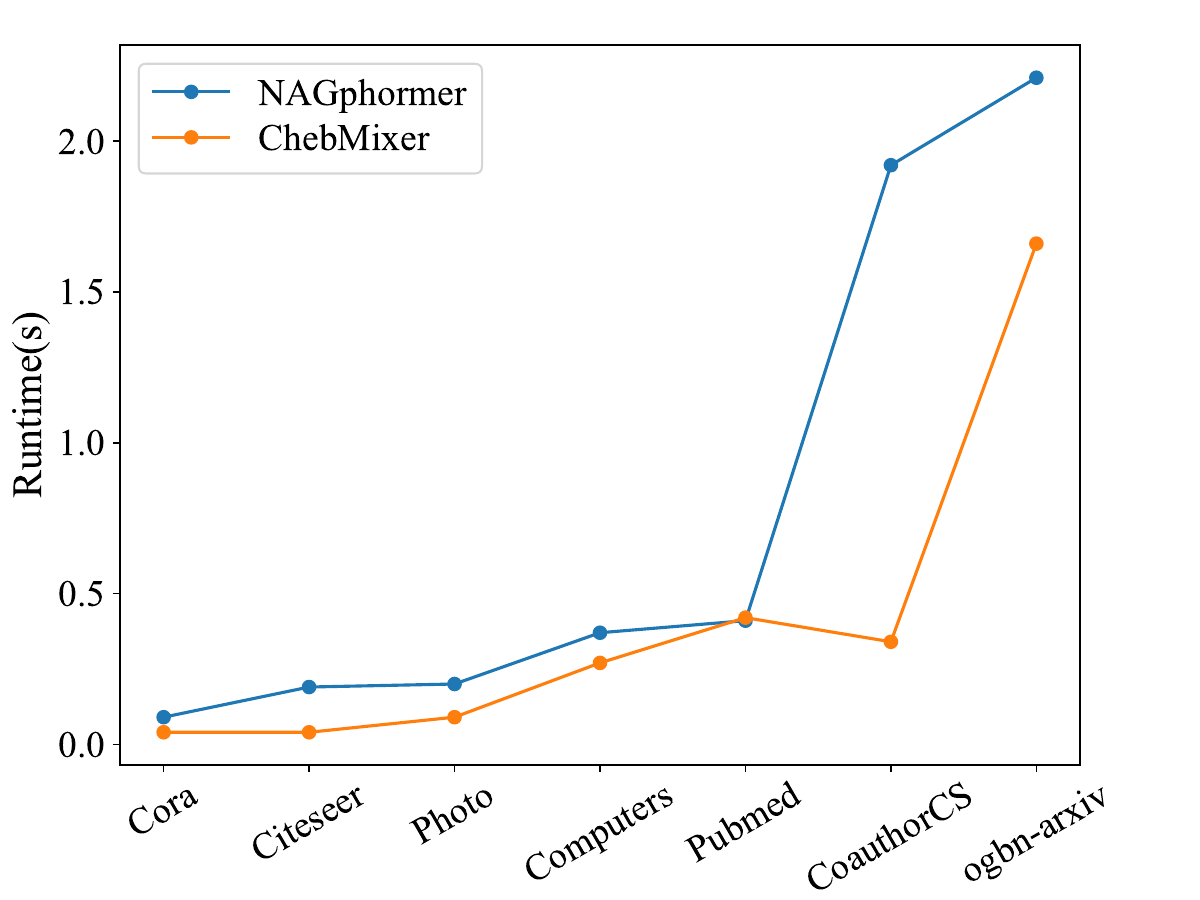}
   \caption{Comparison of computational efficiency with NAGphormer.}
   \label{fig:runtime}
\end{figure}

To evaluate the time efficacy of our proposed method, we conduct a comparative analysis with NAGphormer~\cite{chennagphormer} in terms of epoch-wise training time. For all datasets, the hyperparameter $K$, the hidden dimension $d$, and the number of model layers $l$ are set to 7, 64, and 1, respectively. The experimental results are illustrated in \cref{fig:runtime}. As the dataset size increases, the training time proportionally increases. Notably, our method surpasses NAGphormer in terms of computational efficiency on almost all datasets.

\section{Conclusion}

We propose ChebMixer, a novel and efficient graph MLP mixer for learning graph representation that addresses the issues of the standard MP-GNN. By using fast Chebyshev polynomials spectral filtering to extract multiscale or multi-hop representations of graph nodes, we can treat each node as a sequence of tokens and efficiently enhance different hop information via MLP mixer. After a well-designed aggregator, we can produce more informative node representations to improve the performance of downstream tasks. Experiment results demonstrate that our approach yields state-of-the-art results in graph node classification and medical image segmentation. We hope our attempt at tasks on the different domains will encourage the unified modeling of graph and image signals via graph representation learning.


{
    \small
    \bibliographystyle{ieeenat_fullname}
    \bibliography{main}
}


\end{document}